\documentclass[conference]{IEEEtran}
\IEEEoverridecommandlockouts
\usepackage{cite}
\usepackage{amsmath,amssymb,amsfonts}
\usepackage{algorithmic}
\usepackage{graphicx}
\usepackage{textcomp}
\usepackage{xcolor}
\usepackage{caption}
\usepackage{subcaption}
\usepackage{enumitem}

\def\BibTeX{{\rm B\kern-.05em{\sc i\kern-.025em b}\kern-.08em
    T\kern-.1667em\lower.7ex\hbox{E}\kern-.125emX}}
\usepackage{multicol}
\usepackage[pagebackref=true,breaklinks=true,colorlinks,bookmarks=false]{hyperref}
\begin{document}

\title{Presentation Attack Detection with Advanced CNN Models for Noncontact-based Fingerprint Systems\\
}
\makeatletter
\newcommand{\linebreakand}{%
  \end{@IEEEauthorhalign}
  \hfill\mbox{}\par
  \mbox{}\hfill\begin{@IEEEauthorhalign}
}
\makeatother

\author{\IEEEauthorblockN{Sandip Purnapatra}
\IEEEauthorblockA{\textit{ECE Department} \\
\textit{Clarkson University}\\
Potsdam, USA \\
purnaps@clarkson.edu}
\and
\IEEEauthorblockN{Conor Miller-Lynch}
\IEEEauthorblockA{\textit{CS Department} \\
\textit{Clarkson University}\\
Potsdam, USA \\
millerlc@clarkson.edu}
\and
\IEEEauthorblockN{Stephen Miner}
\IEEEauthorblockA{\textit{CS Department} \\
\textit{Clarkson University}\\
Potsdam, USA \\
minersj@clarkson.edu}
\and
\IEEEauthorblockN{Yu Liu}
\IEEEauthorblockA{\textit{CS Department} \\
\textit{Clarkson University}\\
Potsdam, USA \\
liuy5@clarkson.edu}
\and
\IEEEauthorblockN{Keivan Bahmani}
\IEEEauthorblockA{\textit{ECE Department} \\
\textit{Clarkson University}\\
Potsdam, USA \\
bahmank@clarkson.edu}
\linebreakand
\IEEEauthorblockN{Soumyabrata Dey}
\IEEEauthorblockA{\textit{CS Department} \\
\textit{Clarkson University}\\
Potsdam, USA \\
sdey@clarkson.edu}
\and
\IEEEauthorblockN{Stephanie Schuckers}
\IEEEauthorblockA{\textit{ECE Department} \\
\textit{Clarkson University}\\
Potsdam, USA \\
sschucke@clarkson.edu}
}
\maketitle

\begin{abstract}
Touch-based fingerprint biometrics is one of the most popular biometric modalities with applications in several fields. Problems associated with touch-based techniques such as the presence of latent fingerprints and hygiene issues due to many people touching the same surface motivated the community to look for non-contact-based solutions. For the last few years, contactless fingerprint systems are on the rise and in demand because of the ability to turn any device with a camera into a fingerprint reader. Yet, before we can fully utilize the benefit of noncontact-based methods, the biometric community needs to resolve a few concerns such as the resiliency of the system against presentation attacks. One of the major obstacles is the limited publicly available data sets with inadequate spoof and live data. In this publication, we have developed a Presentation attack detection (PAD) dataset of more than 7500 four-finger images and more than 14,000 manually segmented single-fingertip images, and 10,000 synthetic fingertips (deepfakes). The PAD dataset was collected from six different Presentation Attack Instruments (PAI) of three different difficulty levels according to FIDO protocols, with five different types of PAI materials, and different smartphone cameras with manual focusing. We have utilized DenseNet-121 and NasNetMobile models and our proposed dataset to develop PAD algorithms and achieved PAD accuracy of Attack presentation classification error rate (APCER) 0.14\% and Bonafide presentation classification error rate (BPCER) 0.18\%. We have also reported the test results of the models against unseen spoof types to replicate uncertain real-world testing scenarios.
\end{abstract}

\begin{IEEEkeywords}
Deepfakes, PAD, Finger photo, DenseNet, Noncontact Fingerprint, Contactless Fingerprint, Touchless Fingerprint, Finger photo PAD
\end{IEEEkeywords}

\section{Introduction}
Biometrics is a substitute for inconvenient and insecure password/PIN authentication. The fingerprint is a biometric modality that offers a high level of accuracy, universality, uniqueness, and permanence. Thus, making it a popular and widely used biometric modality by industries, law enforcement agencies, and national ID programs worldwide \cite{lin2018matching} \cite{jain2012biometric}. However, collecting fingerprint data from subjects require various kinds of touch-based sensors, among them the most popular ones are optical, capacitive, and ultrasound. Recent research has revealed that collecting fingerprint data from these sensors can cause various performance issues \cite{labati2015toward}:

\begin{itemize}[noitemsep,leftmargin=*]
\item {\bf Sensors are hard to clean} - Latent fingerprints of a subject might interfere with the following captures of fingerprint
\item {\bf Distortions in fingers for touching sensor surface} - Elastic deformation caused by the friction between finger skin and sensor surface during fingerprint collection can result in reduced performance
\item {\bf Fingerprint capture problems because of certain challenges} - Skin deformation, humidity can lead to low capture contrast and the failure-to-acquire of fingerprints
\item {\bf No universal sensor} - Different types of fingerprint sensors available worldwide and most are used as an accessory to a device. The inclusion of a secure fingerprint sensor in a smartphone raises the cost of the product.
\end{itemize}

Apart from these, due to the global COVID-19 pandemic, another concern with touch-based fingerprint sensors was hygiene and sterilization after each use \cite{grosz2021c2cl}. So, the popularity of touchless or noncontact-based fingerprint sensors has sky-rocketed. One of the main reasons is any device with a camera can potentially be used as a noncontact-based fingerprint sensor, making the use of accessory fingerprint sensors obsolete. In this decade a large volume of banking and e-commerce applications revolve around smartphones and mobile devices, where users access and store sensitive and confidential information. Most high-end smartphones nowadays use biometric sensors to verify authorized users. Presently, smartphone-based fingerprint recognition systems can be generally classified into two different categories: (1) fingerprint-based authentication and (2) finger photo-based recognition \cite{sankaran2015smartphone}. The popularity boom of smartphone usage in the last decade and modern technological advances in smartphones opened a great opportunity for the universal use of smartphones as a noncontact-based fingerprint sensor \cite{sankaran2015smartphone}. A smartphone camera can capture finger-photo and associated software can extract the necessary information from the photos to use it as an authentication technology. However, PAIs or spoofs can cause a security risk to the noncontact-based fingerprint systems and cause a breach of confidentiality or sensitive data leakage \cite{fujio2018face} \cite{marasco2021fingerphoto}. So, it would be important to determine their ability to detect liveness.  

The most significant contributions of this publication are: 

\begin{itemize}[noitemsep,leftmargin=*]
\item {\bf Advanced Convolutional Neural Network (CNN) based PAD algorithm architectures}, tested against unseen PAIs to improve on the state-of-the-art in noncontact-fingerprint PAD. 

\item {\bf A live dataset collected from 35 subjects using 12 different smartphones of 6 different varieties}, using a commercial noncontact fingerprint collection application developed for both android and iPhones.

\item Development of PAD dataset using {\bf five different PAI materials} in accordance to Fast ID Online (FIDO) Biometric Requirements - {\bf six different PAIs} of three difficulty levels (based on time, expertise, and equipment) according to FIDO criterion \cite{schuckers2021fido}. All algorithms were evaluated by standard PAD metrics as defined by International Organization for Standardization (ISO) \cite{ISO_IEC_301073:2017}. Researchers would be able to gain access to this dataset by April 2023 after agreeing to the dataset release agreement as per IRB-approved protocol.

\item Introduction of single-fingertip-based {\bf Synthetic fingertips}: high-quality 10,000 synthetically generated samples from 35 live subjects by using StyleGAN with Adaptive Discriminator Augmentation (ADA) model \cite{karras2020training}.
\end{itemize}
 
The rest of the publication is structured as follows. Section II presents an analysis of the state-of-the-art in noncontact fingerprint PAD. Section III details of the dataset and image processing information. Section IV evaluates the PAD results. Finally, Section V discusses the limitations and concludes the publication.

\section{State-of-the-art}
In the last two decades, researchers have been exploring smartphone-based finger photo recognition \cite{lee2006preprocessing} \cite{lee2008recognizable} \cite{li2012testing} \cite{sankaran2015smartphone}. However, these publications were focused on finger photo-processing techniques i.e. segmentation, enhancement, and matching with the minutia-based algorithms. These algorithms demonstrated the challenges of finger photo processing and the datasets were not made available publicly to encourage further research. In the last decade, Stein et al. \cite{stein2013video} introduced a PAD algorithm for spoof detection in a smartphone-based noncontact fingerprint-capturing application. The algorithm used reflection properties as hand-crafted features to analyze the frames of a finger video to achieve a 77\% successful detection rate, assuming the PAIs i.e. finger photo print out or finger sleeves would have different reflection properties than live fingers. However, from the results, it is not clear if finger sleeves laid on top of live fingers would be accurately detected using this method. Taneja et al. \cite{taneja2016fingerphoto} studied finger photo PAD on mobile devices with print-out and replay attacks and successfully classified between the live and spoof samples using a Support Vector Machine (SVM) algorithm with hand-crafted texture-based features and resulting in a detection equal error rate (D-EER) of 3.71\%. The PAD dataset is publicly available, however, the manual capture of live and finger photos was not properly focused and thus of low quality and noisy, and neither data was collected using standard international collection protocols. Fujio et al. \cite{fujio2018face} using the same dataset developed a PAD algorithm with CNN (AlexNet) which detected the noise of the spoof images, and achieved a half total error rate of 0.04\%. The authors resized the images to 256*256 dimensions for analysis, which raise the chance of loss of further features in the unfocused dataset and might not provide a good classification against high-quality spoofs. The same PAD dataset, along with collected replay attack data, was used by Marasco et al. \cite{marasco2021fingerphoto} to achieve D-EER of 2.14\% for AlexNet and 0.97\% with ResNet. The authors evaluated different color spaces and CNN architectures, however, the publication did not mention the details of the image quality used for analysis. Wasnik et al. \cite{wasnik2018presentation} published PAD-based finger photo recognition. The authors converted the finger photos and replay attacks to traditional digital fingerprints and derived hand-crafted features like second-order local structures and classified between the live and spoof images using an SVM algorithm to achieve 4.43\% D-EER. \par 

Recently, Kolberg et al. \cite{kolberg2023colfispoof} introduced the {\bf COLFISPOOF} dataset for non-contact fingerprint PAD. The dataset has 7200 samples of 72 different PAI species and was captured using two different smartphones. We applaud their effort to collect such large-scale data for deep neural network-based training. Most notably, the PAI materials used by the authors are mostly silicon-based and of FIDO level A \& B difficulty, which translates to easy and moderately difficult to make. Furthermore, the authors used {\bf SynCoLFinGer} \cite{priesnitz2022syncolfinger}, which synthetically simulates and generates finger photos from contact-based fingerprints. These synthetic live fingerprints can be very easily visually distinguished from the live fingertips. The authors also proposed evaluation protocols to train and test the PAD algorithms using the developed dataset but did not contribute any PAD algorithm. \par
From the evaluation of the state-of-the-art, evidently, none of the studies that have shared data sets publicly have a large spectrum of PAIs according to FIDO or ISO standards. The datasets we have found in the public domain are low quality, not in focus, only used finger sleeves or PAIs made artificially from different materials to lay on fingertips to obscure real fingertips, print out of finger photos, and display attacks as PAIs, and none of them used extremely sophisticated, hard to make, and visually indistinguishable synthetic PAIs. Many works have resized their images, without exploring if the resized images were losing important features. The algorithms used for PAD are either hand-crafted feature-based or used older CNN architectures. Neither the algorithms were tested against unseen or unknown spoof varieties to replicate real-life uncertain scenarios. Keeping these in mind, we have developed a single fingertip-based spoof dataset according to the standard PAI creation protocols, and difficulty levels, using different types of materials and spoof textures that reflect real skin tones.  We have also included high-quality synthetically generated fingerprints. All the spoof images were collected manually by focusing on the target fingertip and quality was checked to ensure the fingertip is in focus and of good quality i.e. not blurred or smudged and fingertip ridges and valleys are clearly visible. We have used more recent and advanced CNN architectures i.e. DenseNet-121 and NasNetMobile for developing PAD algorithms and compared the algorithms against similar architecture adoption of Keras \cite{Keras_API_Reference} and tested them against unseen spoofs i.e. our algorithms were not trained with the unseen PAIs to replicate the real-life uncertain scenarios. The model architectures were chosen based on their size, parameters, and performance against the ImageNet validation dataset. All the algorithms used in this publication were used for binary class i.e. live vs. spoof classification.

\section{Dataset and Image processing}
We have constructed a Presentation Attack (PA) dataset of 35 subjects, consisting of two sessions of live data collection and finger-mold collection in the first session for the four fingers of both the subject's hands. The live data was collected using 6 smartphone pairs. The finger mold (made from a dental mold and impression material) was used to create finger molds or PAIs using five different types of materials, additives to achieve human skin tones with ecoflex PAI, and deepfake fingertips or synthetic fingers. Please refer to {\bf Fig: \ref{Example_images}} for visual single-fingertip-based finger photo examples. Table~1 describes the number of images we have presently in our dataset. All the four-finger-based live images were automatically captured using Veridium TouchlessID \cite{Touchless_Veridium}, a commercial smartphone-based noncontact-based fingerprint recognition technology, which provides us with the captured four-finger images of the subject's both hands. The app automatically sets the focus and distance of the captures (we do not have this proprietary information from Veridium). However, we could not use the same app to capture PAIs, because of the failure-to-capture of many PAI images and license issues. Thus we captured the spoof images from the PAI samples overlaid on the real fingertips and using the manual capture with a smartphone camera. During capture, the collectors set the focus of the camera on each fingertip of the hand in a four-finger image capture setting, from a 6-inch distance from the camera. After we finished collecting ecoflex PAIs, we noticed that the collectors were having difficulty capturing good quality, well-focused images from the older smartphones i.e. Samsung Galaxy S6s, S7s, and Google pixels. So, we did not use these phones for the collection of other PAI modalities and only used the most recent smartphones, i.e., iPhone 7, iPhone X, and Samsung Galaxy S20. It would also be important to mention that the collection of PAI images from different PAI modalities is still ongoing. Presently, our dataset has more than 23,000 single fingertip images ref{Dataset}. We plan to collect a further 6000 single-fingertip PAI images by April 2023.
\subsection{Presentation Attack Instrument Generation}
The PAIs were prepared using five different types of materials and collected using different smartphones. Additional details about each PAI species are provided as follows:

\begin{itemize}[noitemsep,leftmargin=*]
    \item {\bf Ecoflex layover (EL):} Fingertip layover molds for each finger of the subject's hand mold were prepared using the ecoflex with makeup additives to achieve colors close to human skin tones and used as PAI of difficulty level B. Ecoflex PAI images of 24 subjects were collected.
    \item {\bf Playdoh layover (PL):} Fingertip layover molds for each finger of the subject's hand mold were prepared using playdoh material of red color (easily available in markets) and used as a PAI of difficulty level B. Playdoh PAI images of 35 subjects were collected.
    \item {\bf Wood glue layover (WL):} Fingertip layover molds for each finger of the subject's hand mold were prepared using wood glue. Wood glue PAI images of all 35 subjects were collected and used as PAI of difficulty level B.
    \item {\bf Synthetic fingertip (SF):} We have manually segmented more than 8100 single-fingertips from the live four-finger images. We have checked all the live images for blur. More details about blur detection are discussed in \ref{Data processing}. After removing the blurred images, we trained the StyleGAN with ADA \cite{karras2020training}. The model requires fewer data for training than the previous StyleGAN models because of ADA. We used the PyTorch version of the original model shared by the authors and generated 10,000 synthetic fingertip images with an 8.25 Frechet Inception Distance (FID) score with a fixed resolution of 512 heights. The synthetic PAIs were used as the PAI of difficulty level C.
    \item {\bf Latex layover (LL):} Fingertip layover molds for each finger of the subject's hand mold were prepared using latex, a rubberized liquid material available in the market for preparing props and prop molds. Latex PAI images of 6 subjects were collected and used as an unseen PAI of difficulty level B to test the performance of our algorithm models. The collection is still ongoing.
    \item {\bf Printed PAI (PP):} Live fingerprint images were printed out on glossy photo paper using a color-jet printer. This PAI modality has been used as an unseen PAI of level A difficulty to test the performance of our algorithms. PAI images of all 35 subjects were collected.
\end{itemize}

\begin{figure*}[htbp]
\centering

\subcaptionbox{\centering Live Single-fingertip image}{\includegraphics[height=3.5cm,width=0.13\textwidth]{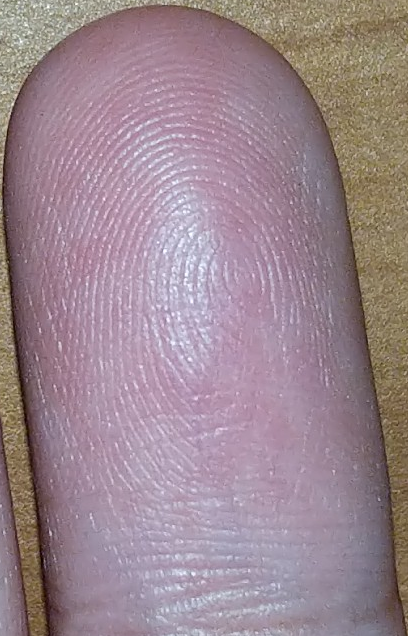}}
\hfill
\subcaptionbox{\centering Ecoflex PAI}{\includegraphics[height=3.5cm,width=0.13\textwidth]{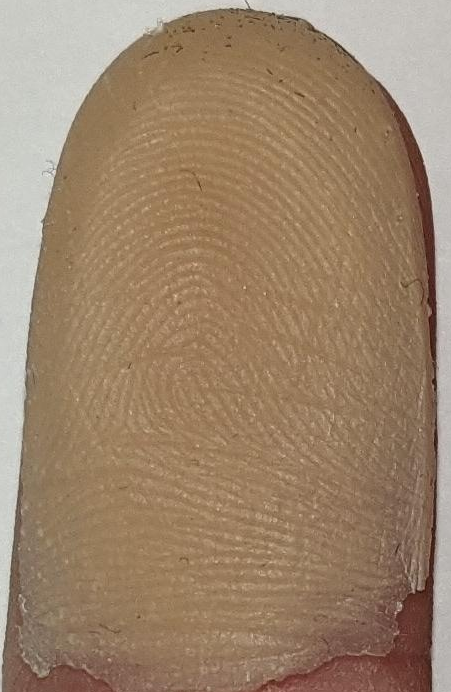}}
\hfill 
\subcaptionbox{\centering Playdoh PAI}{\includegraphics[height=3.5cm,width=0.13\textwidth]{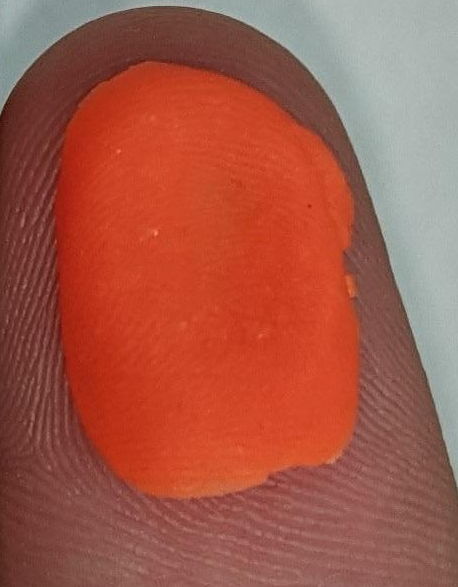}}
\hfill
\subcaptionbox{\centering Wood Glue PAI}{\includegraphics[height=3.5cm,width=0.13\textwidth]{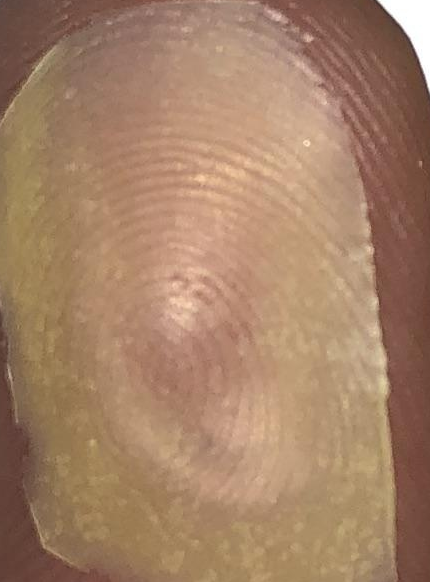}}
\hfill
\subcaptionbox{\centering Latex PAI}{\includegraphics[height=3.5cm,width=0.13\textwidth]{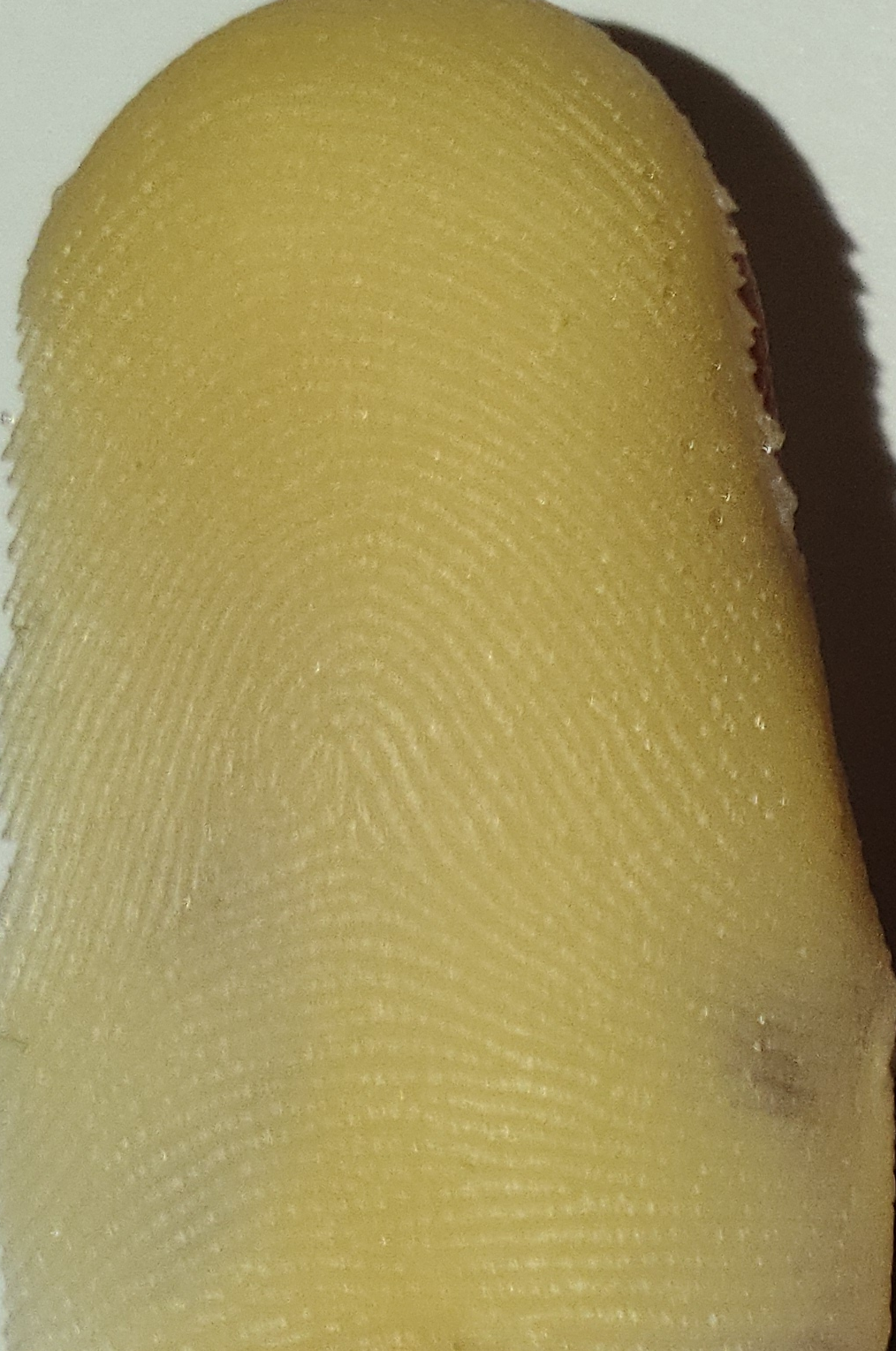}}
\hfill
\subcaptionbox{\centering Glossy Printed Finger Photo PAI}{\includegraphics[height=3.5cm,width=0.13\textwidth]{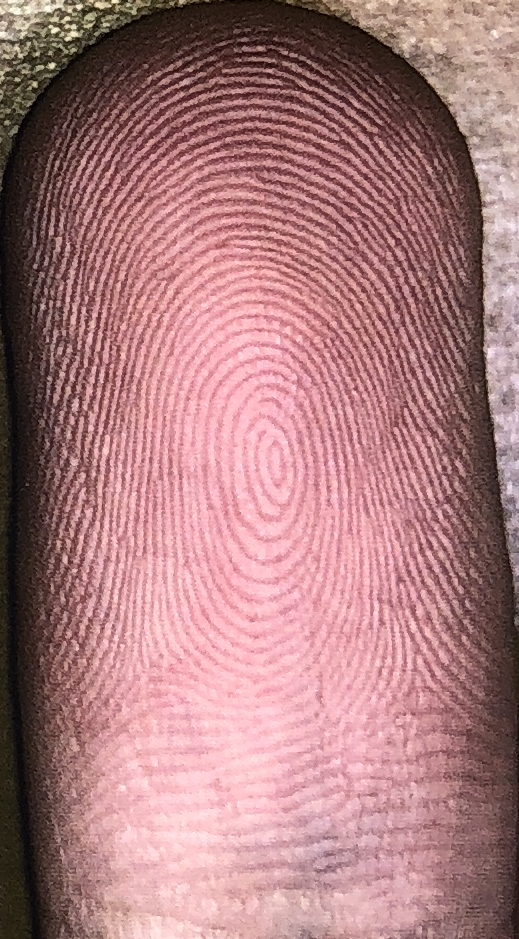}}
\hfill
\subcaptionbox{\centering Synthetic PAI}{\includegraphics[height=3.5cm,width=0.13\textwidth]{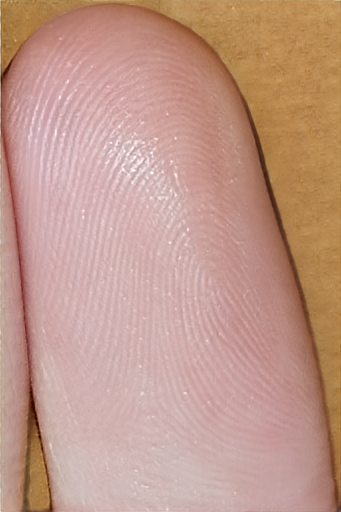}}
\hfill
\caption{Example of all live, PAIs images}
\label{Example_images}
\end{figure*}

Due to the manual segmentation of live and of the PAI dataset (except synthetically generated), the resulting image sizes were variable. After plotting the distribution of the images, we selected a value close to the median resolution to down-sample the images by 20\% to bring them closer to the median. Please refer to Fig: \ref{Distribution} for the image distribution. The same rule was applied to single-finger-based wood glue, finger photo, and latex PAI images. 

\begin{figure*}[htbp]
\centering

\subcaptionbox{\centering Live Single-fingertip image distribution}{\includegraphics[height=4cm,width=0.32\textwidth]{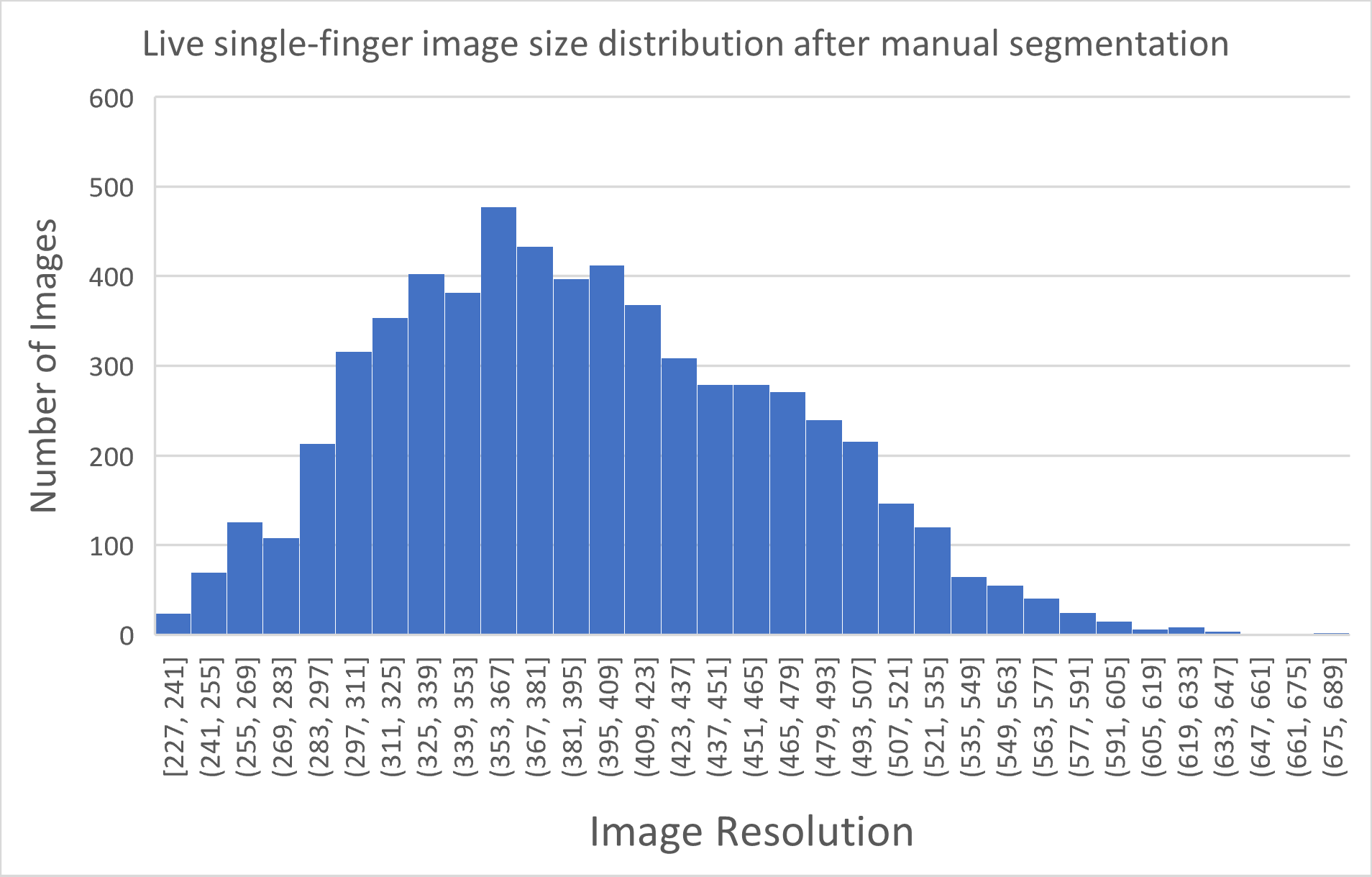}}
\hfill
\subcaptionbox{\centering Ecoflex PAI Single-fingertip image distribution}{\includegraphics[height=4cm,width=0.32\textwidth]{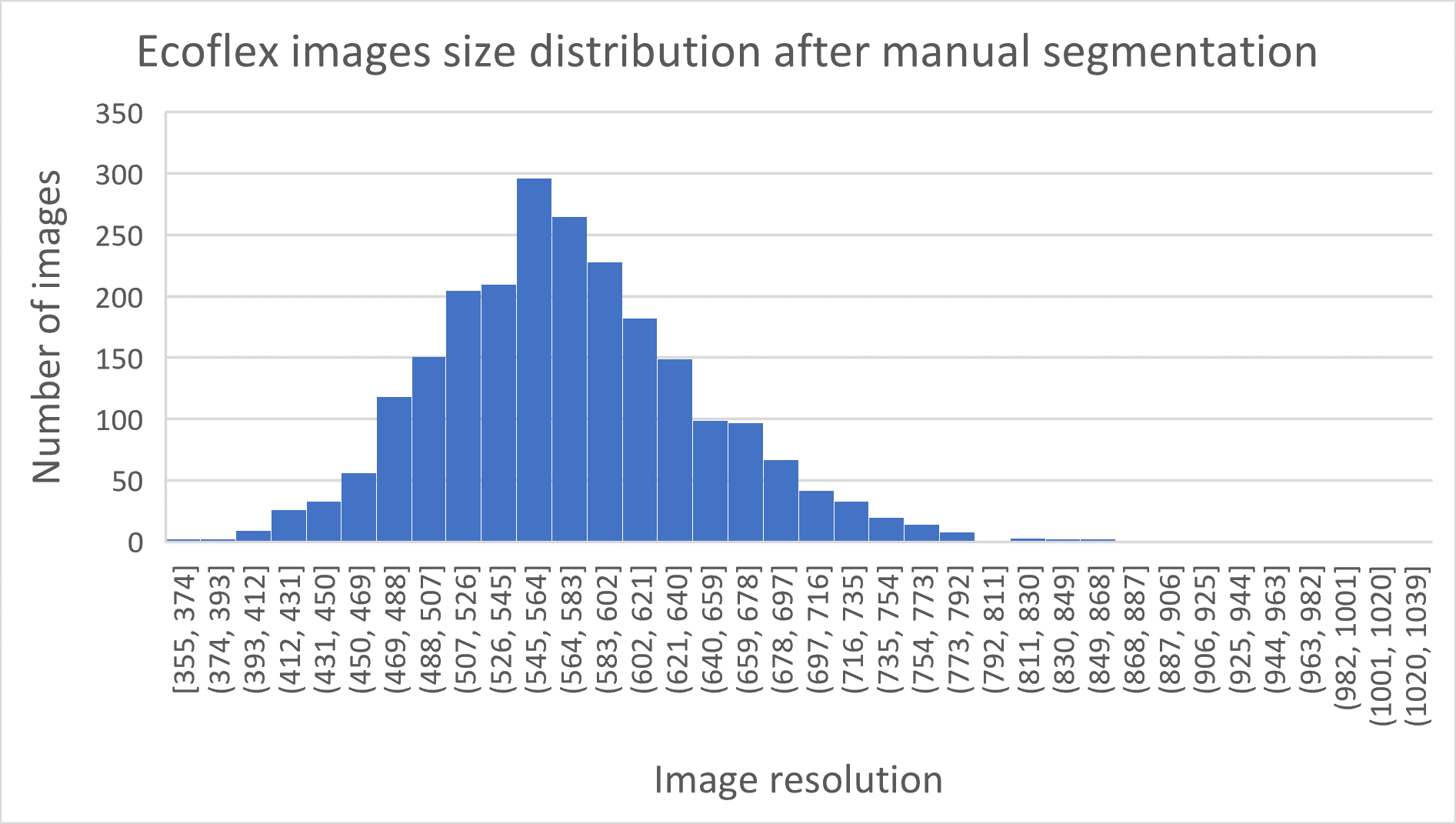}}
\hfill
\subcaptionbox{\centering Playdoh PAI Single-fingertip image distribution}{\includegraphics[height=4cm,width=0.33\textwidth]{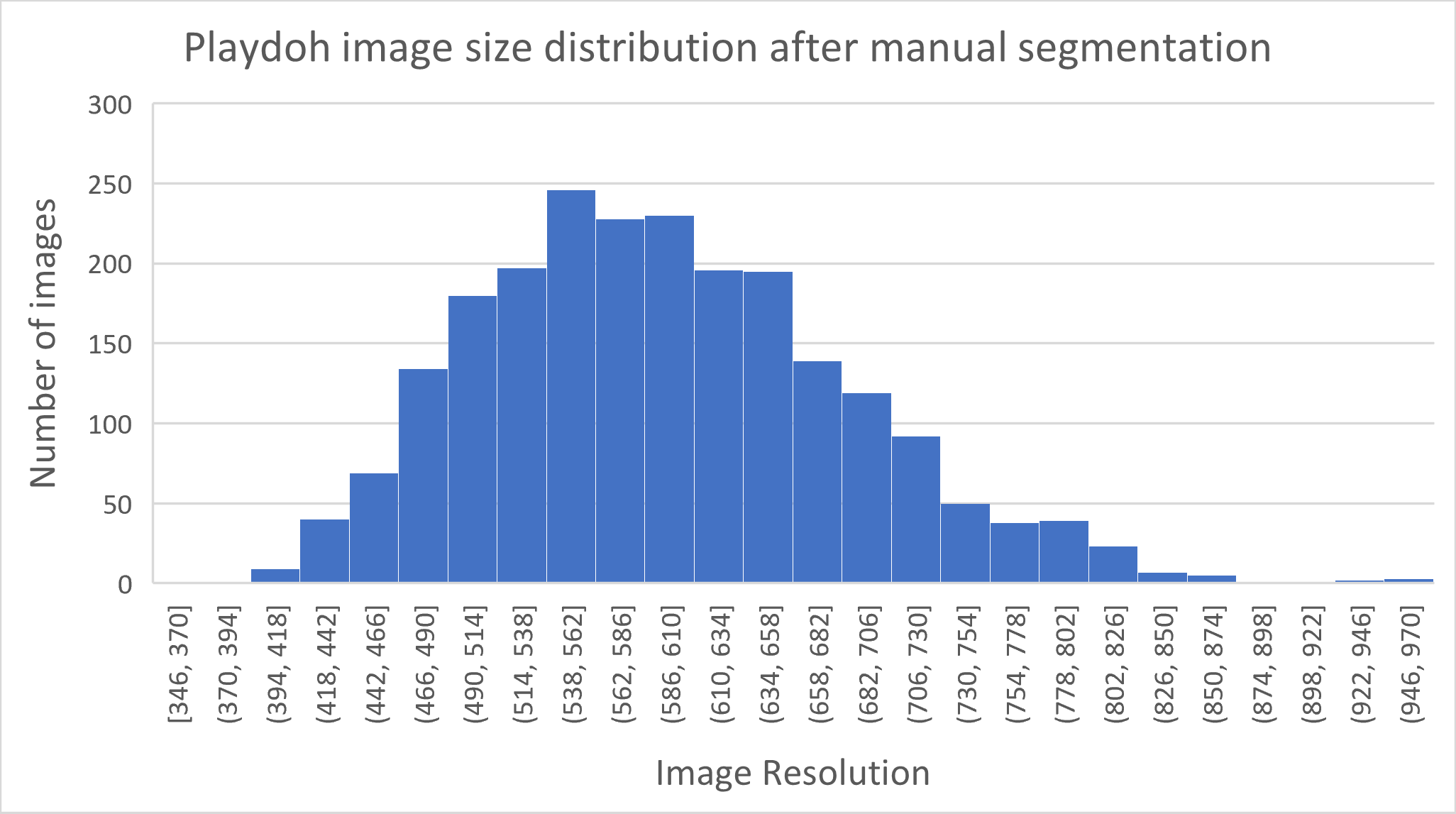}}
\caption{Distribution of all live, PAIs images after manual segmentation}
\label{Distribution}
\end{figure*}

To keep the number of images for each live and PAI class to train our models coherently, we have selected a different number of random 256* 256 image patches from the middle of the image of the live and PAI images. Please refer to Table \ref{Dataset} for details about the number of patches extracted. 

\begin{table*}[!t]
\footnotesize
\centering
\caption{Noncontact Fingerprint PAD Dataset Summary}
\label{Dataset}
\begin{tabular}{|c|c|c|c|c|c|c|c|}
\hline
\textbf{Image type} & \textbf{Four-finger }  & \textbf{Single-fingertips} & \textbf{Random} &\textbf{Total image } &\textbf{Training} &\textbf{Testing} &\textbf{Sensors} \\

& \textbf{images} & \textbf{extracted} & \textbf{patches} & \textbf{patches} & \textbf{Patches} & \textbf{Patches} & \\

\hline
Live & 2150 & 7768 & 4 & 31,702 & 27,556 & 1648 & iPhone X*2, iPhone7*2, \\

 &  & & & &  &  &  Samsung Galaxy S9*2, Google Pixel*2, \\

  &  & & & &  &  &  Samsung Galaxy S6*2 and S7 *2 \\

\hline
Ecoflex PAI & 3684 & 2194 & 4 & 8776 & 7404 & 620 & iPhone X*2, iPhone 7*2 \\

 &  & & & &  &  &  Samsung Galaxy S9*2, S20 \\
 
\hline 
Playdoh PAI & 1672 & 2019  & 4  & 8076 & 6652 & 708 & iPhone X*2, iPhone 7*2 \\

 &  & & & &  &  &  Samsung Galaxy S9*2, S20 \\
 
\hline
Wood Glue PAI & 1122 & 1126 & 7 & 7882 & 6573 & 637 & iPhone X*2, iPhone 7*2  \\
 &  & & & &  &  &  Samsung Galaxy S20\\

\hline
Synthetic PAI & 10000 & 8940  & 1 & 8940 & 7439 & 751 & Same as live image  \\

 & (single-fingertip) & & & &  &  &  Samsung Galaxy S9, Google Pixel \\

\hline
Finger Photo PAI & 1022 & 493 & 2 & 986 & 0 & 986 & iPhone X*2, Samsung Galaxy S20 \\

\hline 
Latex PAI & 48 & 48 & 5 & 240 & 0 & 240 & Samsung Galaxy S20 \\

\hline

\multicolumn{8}{c}{Printed finger Photo and Latex PAI have been used as unknown PAIs for analysis}
\end{tabular}
\end{table*}

\subsection{Image-processing}\label{Data processing}
The collected live and spoof four-finger image data sets were segmented to extract the fingertip regions. We used the Clarkson graphical user interface for manual segmentation, which is based on labelImg, an open-sourced python package \cite{murshed2021deep}. The software reads an image and allows for manual segmentation by manually drawing boundary boxes around an area of interest in the image with a manual labeling option and saves the coordinates of the boundary box. The reason for single fingertip extraction was to generate as many images as we can for the data-hungry advanced CNN models and to ensure resolution for CNN models was of sufficient amount to maintain fingerprint features. More details about the models used are described in section-\ref{algorithms}. However, the number of single-fingertip images was not enough to successfully train our deep algorithm models, so we have extracted a number of random 256*256 patches from the middle of the single-finger images. We are still in the process of collecting and segmenting more live and PAI images, which we hope to share with the researchers when we make this PAD dataset public. \par

During our analysis, we noticed the automatically collected live four-finger images had some blur in some of the fingertips. we decided to check the levels of blur and choose a threshold to discard some of the extracted live single-finger images from the final live dataset. This was primarily done to generate better-quality synthetic fingertips from the StyleGAN-ADA model. We checked each live single-fingertip image and calculated the variance of Laplacian. From those scores, we selected a threshold of 220 that removed 9.8\% of the images we had in our dataset, reference Fig:\ref{fig}. The remaining final data were used for all further analysis.

\begin{figure}[htbp]
\centerline{\includegraphics[height=5cm,width=0.5\textwidth]{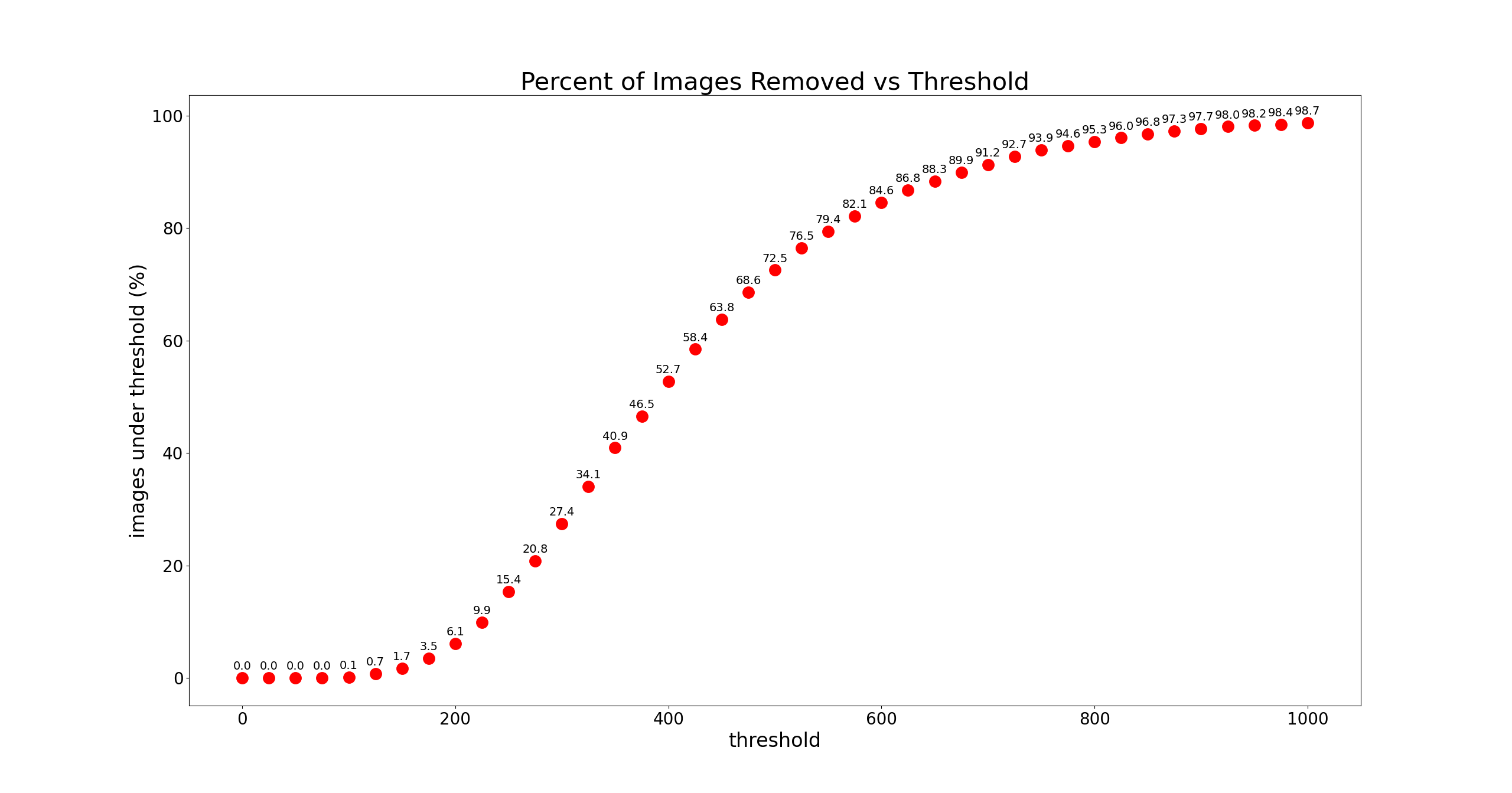}}
\caption{Plot of the percentage of images below a blur threshold}
\label{fig}
\end{figure}

\section{Algorithms and PAD results}\label{algorithms}
In this publication, a variety of deep algorithm architectures, DenseNet-121, DenseNet-121 Keras, and NasNet mobile Keras versions \cite{huang2017densely} \cite{Keras_API_Reference} have been used. The original version of the densely connected convolutional neural network paper \cite{huang2017densely}, shared the code in .LUA version. We have manually converted the DenseNet with 121 layers (with a fixed growth rate of 12) to .PY to use it for our experiments. One of the main reasons for choosing this model as our PAD algorithm architecture is that it is an advanced ResNet architecture that is computationally less expensive. The dense blocks used in the DenseNet improve the feature learning process by leveraging a transition layer i.e. the combination of convolution, average pooling, and batch normalization between two dense blocks. This helps concatenate feature maps and reduce the vanishing gradient problems. The gradients flow from the initial input layer and loss function are shared with all the layers. Thus reducing the number of required parameters, and features, providing a much less computationally expensive model compared to other advanced CNN model architectures i.e. AlexNet, ResNet, VGG16, and VGG19. We have compared our converted version of DenseNet against the standard DenseNet architecture (with a fixed growth rate of 64) developed by Keras \cite{Keras_API_Reference}. In terms of performance against known PAIs and computational expense, our model significantly outperforms the Keras version of DenseNet. One important point needs to be mentioned that in the Keras version of DenseNet, we have used the ImageNet pre-trained weights for pre-training, and because of that, the input image size of the model was $224\times224$ dimension. Due to the image size difference, the input images of the Keras version of the DenseNet were resized from $256\times256$ to $224\times224$. We have also used the Keras version of NasNetMobile to check the performance of the NasNetMobile model against our dataset. NasNetMobile has close top-1 and top-5 percent accuracy as DenseNet, on the ImageNet validation dataset but it is less computationally expensive than the Keras version of the DenseNet. NasNet mobile model also used ImageNet pre-trained weights and the input images of 256*256 dimension were resized to 224*224 dimension to fit the model. We used common augmentations like the rotation of 45 degrees, horizontal flip, and a zoom range of 0.1 for all of our models. Please refer to Table \ref{result} for model performance comparison details.

We calculated the APCER and BPCER of each PAI Species and live data of our test dataset. From the results, we can categorize the model performances in two different categories i.e. performance against known PAIs or the PAIs the models were trained with and the unknown PAIs or the PAIs which were not used for model training and only used in the test dataset to understand the generalization of the models against unseen spoof types. In terms of the performance of the models against known PAI categories, our converted DenseNet-121 model has outperformed the other two models developed by Keras, with APCER 0.14\%. In terms of performance against live data, our converted DenseNet-121 model outperforms the other two models with BPCER 0.18\%. The Keras version of the DenseNet-121 model outperformed all other models in the unseen PAI categories. The model performed very well against the latex layover PAIs with APCER 0\%, yet, did not perform quite well against color-printed photo paper PAIs with APCER 79.01\%. The NasNetMobile version of Keras did not perform well compared to the Keras DensNet-121 model with APCER 17.08\% against the unknown latex PAIs and APCER 82.15\% against the unknown printed glossy photo paper PAIs. We believe both Keras-developed DenseNet-121 and NasNetMobile performed better against unseen PAIs than our converted DenseNet-121 model, because of the ImageNet pre-trained weights, which provide a better generalization ability of the models against unseen PAIs. 

We tried to find an answer to the research question brought forward by the authors of \cite{kolberg2023colfispoof}: {\bf Can we add color during the PAI casting process to
obtain more threatening skin-colored PAIs?} To find an answer, we used our converted DenseNet model and tested the model with the known ecoflex PAI. The ecoflex PAI is prepared with makeup additives to achieve a close human skin tone. We have trained and tested the performance of the model again, with the dataset converted into grayscale from the RGB scale. We noticed a slight degradation in APCER for ecoflex spoofs and trainable parameters, with all other model parameters remaining unchanged, which does not reflect the importance of skin-colored PAIs. 

\begin{table*}[htbp]
\footnotesize
\centering
\caption{Noncontact Fingerprint PAD Result Summary}
\label{result}
\begin{tabular}{|l|l|l|l|l|l|l|l|l|l|l|l|l|}
\hline
\textbf{Model} & \textbf{Total training}  & \textbf{Image} & \textbf{Trainable} &\textbf{Validation} &\textbf{Test} & \multicolumn{6}{c|}{\textbf{APCER (\%)}}&\textbf{BPCER} \\
\cline{7-12}
 \textbf{Architecture} & \textbf{images} & \textbf{dimension} & \textbf{parameters} & \textbf{images} & \textbf{images} & EL & PL & WL & SF & LL & PP & (\%) \\

\hline
\textbf{DenseNet-121} & 55,624 & 256*256 & 561,002 & 4758 & 5590 & 0 & 0.14 & 0 & 0.13 & 42.5& 88.03 & 0.18\\

\hline
\textbf{DenseNet-121} & 55,624 & 224*224 & 8,619,074 & 4758 & 5590 & 0 & 1.55 & 0.94 & 0.79 & 0 & 79.01 & 3.64 \\
\textbf{(Keras)} & & & & & & & & & & & & \\

\hline
\textbf{NasNetMobile} & 55,624 & 224*224 & 5,884,054 & 4758 & 5590 & 3.22 & 0.71 & 5.96 & 4.12 & 17.08 & 82.15 & 9.04 \\
\textbf{(Keras)} & & & & & & & & & & & & \\

\hline
\textbf{DenseNet-121} & 55,624 & 256*256 & 560,570 & 4758 & 5590 & 0.16 & 1.98 & 11 & 11.85 & 100 & 98.9 & 0.18\\
\textbf{(grayscale)} & & & & & & & & & & & &\\
\hline

\multicolumn{13}{c}{Acronyms: EL: Ecoflex Layover; PL: Playdoh Layover; WL: Woodglue layover; SF: Synthetic Fingertip; LL: Latex Layover; PP: Printed Photo Paper}
\end{tabular}
\end{table*}

\section{Limitations and Discussion}
Noncontact fingerprint based on finger photos is fast becoming popular and has the potential to replace traditional fingerprint authentication methods in several use cases. However, it is important to look into the system's resilience against PAs to increase the security and reliability of noncontact-based fingerprint systems. The availability of a standard PAD dataset with various PAIs with a good number of reference images for the three difficulty levels, captured with multiple sensors and reference PAD models, would boost research interest in this area. For that purpose, we will be releasing the four-finger-based and single-finger-based PAD datasets to interested researchers. As we continue to collect more images and more PAI modalities, we will share the dataset by the date of the conference. The final dataset would have more than 10,000 four-finger-based PAD images and more than 45,000 single-finger-based PAD images with PAIs of six different materials and difficulty levels. Also, there is further scope to develop 3D printed and etched molds and develop PAIs to reflect uncooperative attacks based on stolen biometric images or latent prints.

We have used advanced CNN architectures like DenseNet and NasNetMobile as PAD models. We achieved the lowest APCER of 0\% and BPCER of 0.18\% against known PAIs and live data with our converted DenseNet-121 model. Also, we achieved the lowest APCER of 0\% against latex PAIs, replicating the unknown real-life scenario using the Keras version of the DenseNet-121 model. Although the algorithm results significantly improve the state-of-the-art, there is still scope for further development. Further PAD result comparison with other advanced CNN models with convolutional scaling like EfficientNet B0-B2 can also be considered for analysis. Even with the introduction of this new dataset, more data is still needed to improve the models' performance and generalization to unknown spoof types.

\section*{Acknowledgment}
This material is based upon the work supported in part by the
National Science Foundation under Grant No. 102505, the Center for Identification Technology and Research (CITeR).

\bibliographystyle{IEEEtran}
\bibliography{references.bib}

\begin{thebibliography}{10}
\providecommand{\url}[1]{#1}
\csname url@samestyle\endcsname
\providecommand{\newblock}{\relax}
\providecommand{\bibinfo}[2]{#2}
\providecommand{\BIBentrySTDinterwordspacing}{\spaceskip=0pt\relax}
\providecommand{\BIBentryALTinterwordstretchfactor}{4}
\providecommand{\BIBentryALTinterwordspacing}{\spaceskip=\fontdimen2\font plus
\BIBentryALTinterwordstretchfactor\fontdimen3\font minus
  \fontdimen4\font\relax}
\providecommand{\BIBforeignlanguage}[2]{{%
\expandafter\ifx\csname l@#1\endcsname\relax
\typeout{** WARNING: IEEEtran.bst: No hyphenation pattern has been}%
\typeout{** loaded for the language `#1'. Using the pattern for}%
\typeout{** the default language instead.}%
\else
\language=\csname l@#1\endcsname
\fi
#2}}
\providecommand{\BIBdecl}{\relax}
\BIBdecl

\bibitem{lin2018matching}
C.~Lin and A.~Kumar, ``Matching contactless and contact-based conventional
  fingerprint images for biometrics identification,'' \emph{IEEE Transactions
  on Image Processing}, pp. 2008--2021, 2018.

\bibitem{jain2012biometric}
A.~K. Jain and A.~Kumar, ``Biometric recognition: an overview,'' \emph{Second
  generation biometrics: The ethical, legal and social context}, 2012.

\bibitem{labati2015toward}
R.~D. Labati, A.~Genovese, V.~Piuri, and F.~Scotti, ``Toward unconstrained
  fingerprint recognition: A fully touchless 3-d system based on two views on
  the move,'' \emph{IEEE transactions on systems, Man, and cybernetics:
  systems}, 2015.

\bibitem{grosz2021c2cl}
S.~A. Grosz, J.~J. Engelsma, E.~Liu, and A.~K. Jain, ``C2cl: Contact to
  contactless fingerprint matching,'' \emph{IEEE Transactions on Information
  Forensics and Security}, 2021.

\bibitem{sankaran2015smartphone}
A.~Sankaran, A.~Malhotra, A.~Mittal, M.~Vatsa, and R.~Singh, ``On smartphone
  camera based fingerphoto authentication,'' in \emph{2015 IEEE 7th
  International Conference on Biometrics Theory, Applications and Systems
  (BTAS)}.\hskip 1em plus 0.5em minus 0.4em\relax IEEE, 2015.

\bibitem{fujio2018face}
M.~Fujio, Y.~Kaga, T.~Murakami, T.~Ohki, and K.~Takahashi, ``Face/fingerphoto
  spoof detection under noisy conditions by using deep convolutional neural
  network.'' in \emph{BIOSIGNALS}, 2018, pp. 54--62.

\bibitem{marasco2021fingerphoto}
E.~Marasco and A.~Vurity, ``Fingerphoto presentation attack detection:
  Generalization in smartphones,'' in \emph{2021 IEEE International Conference
  on Big Data (Big Data)}.\hskip 1em plus 0.5em minus 0.4em\relax IEEE, 2021.

\bibitem{schuckers2021fido}
S.~Schuckers, G.~Cannon, and N.~Tekampe, ``{FIDO Biometrics Requirements},''
  \url{https://fidoalliance.org/specs/biometric/requirements/}, 2021, accessed:
  2023-01-19.

\bibitem{ISO_IEC_301073:2017}
{ISO/IEC 30107-3}, ``{Information technology -- Biometric presentation attack
  detection -- Part 3: Testing and reporting},'' 2016.

\bibitem{karras2020training}
T.~Karras, M.~Aittala, J.~Hellsten, S.~Laine, J.~Lehtinen, and T.~Aila,
  ``Training generative adversarial networks with limited data,''
  \emph{Advances in Neural Information Processing Systems}, 2020.

\bibitem{lee2006preprocessing}
C.~Lee, S.~Lee, J.~Kim, and S.-J. Kim, ``Preprocessing of a fingerprint image
  captured with a mobile camera,'' in \emph{International conference on
  biometrics}.\hskip 1em plus 0.5em minus 0.4em\relax Springer, 2006, pp.
  348--355.

\bibitem{lee2008recognizable}
D.~Lee, K.~Choi, H.~Choi, and J.~Kim, ``Recognizable-image selection for
  fingerprint recognition with a mobile-device camera,'' \emph{IEEE
  Transactions on Systems, Man, and Cybernetics, Part B (Cybernetics)},
  vol.~38, no.~1, pp. 233--243, 2008.

\bibitem{li2012testing}
G.~Li, B.~Yang, R.~Raghavendra, and C.~Busch, ``Testing mobile phone camera
  based fingerprint recognition under real-life scenarios,'' \emph{NISK},
  vol.~1, p.~2, 2012.

\bibitem{stein2013video}
C.~Stein, V.~Bouatou, and C.~Busch, ``Video-based fingerphoto recognition with
  anti-spoofing techniques with smartphone cameras,'' in \emph{2013
  International Conference of the BIOSIG Special Interest Group
  (BIOSIG)}.\hskip 1em plus 0.5em minus 0.4em\relax IEEE, 2013.

\bibitem{taneja2016fingerphoto}
A.~Taneja, A.~Tayal, A.~Malhorta, A.~Sankaran, M.~Vatsa, and R.~Singh,
  ``Fingerphoto spoofing in mobile devices: a preliminary study,'' in
  \emph{2016 IEEE 8th International Conference on Biometrics Theory,
  Applications and Systems (BTAS)}.\hskip 1em plus 0.5em minus 0.4em\relax
  IEEE, 2016.

\bibitem{wasnik2018presentation}
P.~Wasnik, R.~Ramachandra, K.~Raja, and C.~Busch, ``Presentation attack
  detection for smartphone based fingerphoto recognition using second order
  local structures,'' in \emph{2018 14th International Conference on
  Signal-Image Technology \& Internet-Based Systems (SITIS)}.\hskip 1em plus
  0.5em minus 0.4em\relax IEEE, 2018.

\bibitem{kolberg2023colfispoof}
J.~Kolberg, J.~Priesnitz, C.~Rathgeb, and C.~Busch, ``Colfispoof: A new
  database for contactless fingerprint presentation attack detection
  research,'' in \emph{Proceedings of the IEEE/CVF Winter Conference on
  Applications of Computer Vision}, 2023.

\bibitem{priesnitz2022syncolfinger}
J.~Priesnitz, C.~Rathgeb, N.~Buchmann, and C.~Busch, ``Syncolfinger: Synthetic
  contactless fingerprint generator,'' \emph{Pattern Recognition Letters}, vol.
  157, pp. 127--134, 2022.

\bibitem{Keras_API_Reference}
Keras, ``{Keras Applications},'' \url{https://keras.io/api/applications/},
  accessed: 2023-01-18.

\bibitem{Touchless_Veridium}
Veridium, ``{TouchlessID},''
  \url{https://www.veridiumid.com/wp-content/uploads/2021/01/Data-Sheet-Veridium-4-Fingers-Touchless-Id.pdf},
  accessed: 2023-01-18.

\bibitem{murshed2021deep}
M.~S. Murshed, R.~Kline, K.~Bahmani, F.~Hussain, and S.~Schuckers, ``Deep slap
  fingerprint segmentation for juveniles and adults,'' in \emph{2021 IEEE
  International Conference on Consumer Electronics-Asia (ICCE-Asia)}.\hskip 1em
  plus 0.5em minus 0.4em\relax IEEE, 2021, pp. 1--4.

\bibitem{huang2017densely}
G.~Huang, Z.~Liu, L.~Van Der~Maaten, and K.~Q. Weinberger, ``Densely connected
  convolutional networks,'' in \emph{Proceedings of the IEEE conference on
  computer vision and pattern recognition}, 2017.

\end{thebibliography}

\end{document}